\documentclass[letterpaper, 10 pt, conference]{ieeeconf}
\IEEEoverridecommandlockouts
\usepackage{fancyhdr}

\fancypagestyle{preprint}{
  \fancyhf{}
  \fancyhead[L]{PREPRINT VERSION}
  
  \setlength{\topmargin}{-0.7in}
  \setlength{\headheight}{15pt}
  \setlength{\headsep}{25pt}
}

\usepackage{graphicx}
\usepackage{cite}  
\usepackage{amsmath,amssymb}
\usepackage{bm}
\usepackage{dsfont}
\usepackage{tabularx}
\usepackage{array}
\usepackage{makecell}
\usepackage{booktabs}
\usepackage{multirow}
\usepackage{comment}

\usepackage{algorithm}
\usepackage{algorithmicx}
\usepackage{algpseudocode}
\usepackage{wrapfig}
\usepackage{subcaption}

\usepackage{siunitx}
\usepackage{threeparttable}
\usepackage{adjustbox}
\usepackage{xcolor}
\usepackage{pifont}
\newcommand{\cmark}{\textcolor{green!60!black}{\ding{51}}} 
\newcommand{\xmark}{\textcolor{red}{\ding{55}}}            
\renewcommand{\cmark}{\textcolor{green!60!black}{$\checkmark$}}
\renewcommand{\xmark}{\textcolor{red}{$\times$}}

\newcolumntype{N}{>{\centering\arraybackslash}m{1.15cm}}
\newcolumntype{L}[1]{>{\raggedright\arraybackslash}p{#1}}
\newcolumntype{C}[1]{>{\centering\arraybackslash}p{#1}}
\newcolumntype{Y}{>{\raggedright\arraybackslash}X}

\algrenewcommand\algorithmicindent{1.2em}

\usepackage{tabularx}
\usepackage{booktabs}
\usepackage[table]{xcolor}
\usepackage{array}
\usepackage{siunitx}
\usepackage{hyperref}

\newcommand{\algoname}{PioneeR}
\newcommand{\algonamelong}{\textbf{P}ositive and negat\textbf{i}ve dem\textbf{o}nstration de\textbf{n}sity–driv\textbf{e}n r\textbf{e}wards with \textbf{R}ule-based specifications}

\title{Learning Social Navigation from Positive and Negative Demonstrations and Rule-Based Specifications}

\author{Chanwoo Kim$^{1}$, Jihwan Yoon$^{1}$, Hyeonseong Kim$^{1}$, Taemoon Jeong$^{1}$, Changwoo Yoo$^{2}$, Seungbeen Lee$^{3,4}$, \\ Soohwan Byeon$^{5}$, Hoon Chung$^{5}$, Matthew Pan$^{6}$, Jean Oh$^{4}$, Kyungjae Lee$^{8}$, and Sungjoon Choi$^{1}\textsuperscript{*}$%
\thanks{${^1}$Chanwoo Kim, Jihwan Yoon, Hyeonseong Kim, Taemoon Jeong, and Sungjoon Choi are with the Department of Artificial Intelligence, Korea University, Seoul, Republic of Korea. (e-mails: \{chanwoo-kim, yoonmungchi, hyeonseong-kim, taemoon-jeong, sungjoon-choi\}@korea.ac.kr)}
\thanks{${^2}$Changwoo Yoo is with the Department of Computer Science and Engineering, Seoul, Republic of Korea. (e-mail: cwyoo01@korea.ac.kr)}
\thanks{${^3}$Seungbeen Lee is with the Department of Artificial Intelligence, Yonsei University, Seoul, Republic of Korea. (email: seungblee@yonsei.ac.kr)}
\thanks{$^{^4}$Seungbeen Lee and Jean Oh are with the Robotics Institute, School of Computer Science at Carnegie Mellon University, Pittsburgh, PA, USA. (email: seungbel@andrew.cmu.edu, jeanoh@cs.cmu.edu)}
\thanks{${^5}$Soohwan Byeon and Hoon Chung are with Mobinn, Suwon, Republic of Korea. (e-mail: \{soohwan.byeon, h.chung\}@mobinn.co.kr))}
\thanks{${^6}$Matthew Pan is with the Department of Electrical and
Computer Engineering, Queens University, Kingston, Canada. (e-mail: matthew.pan@queensu.ca)}
\thanks{$^{7}$Kyungjae Lee is with the Department of Statistics, Korea University, Seoul, Republic of Korea. (email: kyungjae\_lee@korea.ac.kr)}%
}

\begin{document}
\maketitle
\thispagestyle{preprint}
\pagestyle{preprint}

\begin{abstract}
Mobile robot navigation in dynamic human environments requires policies that balance adaptability to diverse behaviors with compliance to safety constraints. 
We hypothesize that integrating data-driven rewards with rule-based objectives enables navigation policies to achieve a more effective balance of adaptability and safety. 
To this end, we develop a framework that learns a density-based reward from positive and negative demonstrations and augments it with rule-based objectives for obstacle avoidance and goal reaching. 
A sampling-based lookahead controller produces supervisory actions that are both safe and adaptive, which are subsequently distilled into a compact student policy suitable for real-time operation with uncertainty estimates. 
Experiments in synthetic and elevator co-boarding simulations show consistent gains in success rate and time efficiency over baselines, and real-world demonstrations with human participants confirm the practicality of deployment. 
A video illustrating this work can be found on our project page \url{https://chanwookim971024.github.io/PioneeR/}.
\end{abstract}

\section{Introduction}\label{sec:introduction}
Mobile robot navigation in crowded, human-shared environments is inherently safety-critical and requires policies that remain reliable while adapting to diverse human behaviors. 
Core challenges~\cite{singamaneni2024survey,mavrogiannis2023core} include uncertainty in human intent, variability in motion patterns, dense interactions and bottlenecks, compliance with social conventions and right-of-way, and strict real-time requirements on embedded platforms. 
Addressing these challenges is essential for socially aware and reliable navigation.

Classical approaches~\cite{fox2002dynamic,khatib1986real,van2011reciprocal,ames2019control,xiao2021rule,wang2025safe} provide interpretability and explicit safety guarantees but often rely on carefully specified objectives and handcrafted rules, making them difficult to generalize in socially dynamic contexts~\cite{singamaneni2024survey}. 
Learning-based methods~\cite{chen2017socially,liu2020robot,xie2021towards,liu2023intention} instead seek to capture human interaction patterns directly from data, enabling adaptive and socially responsive behaviors. 
However, reinforcement learning typically demands extensive reward shaping and large training budgets, while imitation learning is more data-efficient~\cite{pfeiffer2018reinforced} yet remains prone to distributional shifts and lacks explicit safety mechanisms~\cite{ross2011reduction,abbeel2004apprenticeship}. 
These limitations motivate designs that combine the adaptability of learning with the reliability of explicit safety specifications.

We hypothesize that integrating data-driven rewards with rule-based objectives enables navigation policies that achieve a more effective balance of adaptability and safety. 
To this end, we develop a framework that learns a density-based reward map from positive and negative demonstrations and augments it with rule-based objectives for obstacle avoidance and goal reaching. 
A teacher policy evaluates short-horizon rollouts under this formulation, producing supervisory actions that are both adaptive to demonstrated behaviors and explicitly safe by design. 
For real-time deployment, the teacher is distilled into a compact student policy that conditions directly on observations, inheriting adaptability and safety while remaining suitable for embedded operation.

The main contributions of this work are threefold: 
(i) a reward formulation that integrates density-based learning from both positive and negative demonstrations with rule-based specifications for obstacle avoidance and goal reaching;  
(ii) a teacher policy built on this formulation that provides adaptive and safe supervision, together with an uncertainty-aware distillation process that yields a compact policy for real-time operation; and  
(iii) evaluation in elevator co-boarding scenarios, in both simulation and real-world trials, assessing the effectiveness of combining data-driven rewards with rule-based safety in dynamic human environments.

\section{Related Work}
\subsection{Navigation in Socially Dynamic Environments}
Navigation in socially dynamic environments has been approached through classical, learning-based, and hybrid paradigms. 
Classical methods such as window-based search~\cite{fox2002dynamic}, potential fields~\cite{khatib1986real}, velocity–obstacle formulations~\cite{van2011reciprocal}, and control barrier functions~\cite{ames2019control,xiao2021rule,wang2025safe} provide interpretable objectives with explicit safety guarantees. 
However, their reliance on handcrafted rules and cost functions limits scalability in crowded human settings~\cite{singamaneni2024survey}. 
Learning-based methods instead capture interaction patterns directly from data~\cite{chen2017socially,liu2020robot,xie2021towards,liu2023intention}. 
Reinforcement learning~\cite{chen2017socially,liu2020robot,xie2021towards,liu2023intention} can produce socially compliant behaviors but often requires extensive reward shaping and large training budgets, while imitation learning is more data-efficient~\cite{pfeiffer2018reinforced} but remains prone to distributional shifts and lacks explicit safety mechanisms~\cite{ross2011reduction,abbeel2004apprenticeship}. 
To mitigate these limitations, hybrid frameworks integrate learning with rule-based or optimization modules~\cite{bektacs2022apf,dey2023learning,zhu2021rule,long2021learning}, combining adaptability with safety structure, though they often still depend on handcrafted logic and are typically evaluated in simplified environments.

\begin{figure*}[!t]
    \centering
    \includegraphics[width=1.98\columnwidth]{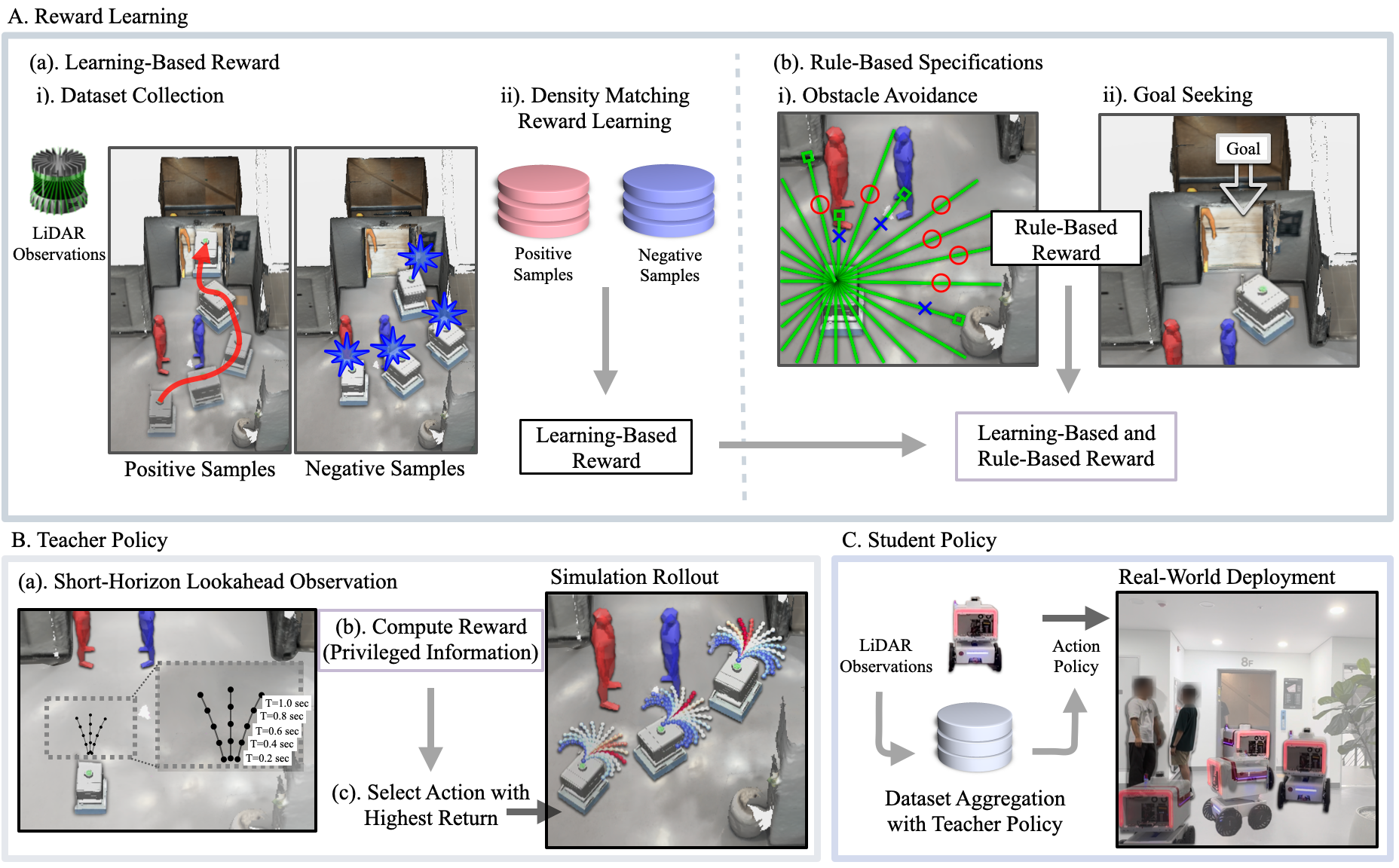}
    \caption{
    Overview of the proposed framework. 
    A. Reward learning: (a) density-based reward maps are constructed from positive and negative demonstrations, and (b) augmented with rule-based specifications for obstacle avoidance and goal reaching. 
    B. Teacher policy: short-horizon candidate rollouts are simulated, scored with the combined reward, and used to select safe and adaptive supervisory actions. 
    C. Student policy: the teacher’s guidance is distilled into a compact policy conditioned on LiDAR observations, enabling real-world deployment. 
    }
    \label{fig:overview}
    \vspace{-4mm}
\end{figure*}

\subsection{Data-Driven Reward Learning}
Another research direction focuses on leveraging demonstrations of varying quality. 
Smooth leveraged kernels~\cite{choi2016robust,choi2019robust} weight demonstrations by quality to improve robustness, while suboptimal or unsafe trajectories serve as counterexamples to delineate undesirable behaviors~\cite{choi2016robust,choi2019robust,oh2022towards}. 
Density-matching reward learning~\cite{choi2016density} further aligns state–action visitation distributions, producing occupancy-like reward maps that emphasize feasible and socially compliant behaviors. 
These advances provide practical means of integrating demonstration-driven learning with explicit safety specifications. 
Inspired by this line of work, we develop a framework that constructs a density-based reward from positive and negative demonstrations and augments it with rule-based terms for obstacle avoidance and goal reaching, enabling navigation that is both safe and adaptive in dynamic human environments.

\section{Problem Formulation}
\subsection{State and Observation}\label{subsec:state_obs}
Let $x\in\mathbb{R}^n$ denote the robot state and $u\in\mathcal{U}\subset\mathbb{R}^m$ the control input. 
The robot dynamics are described by $\dot{x} = f(x,u)$,
and at each decision step the robot issues a velocity command of the form
\begin{equation}
u = [v,\omega]^\top \in \mathcal{U}, \qquad v \in \mathbb{R}_{\ge 0},\;\; \omega \in \mathbb{R},
\end{equation}
where $v$ and $\omega$ denote the translational and rotational velocities, respectively. 
In our case, we assume a unicycle model, i.e., $x = (p_x, p_y, \theta) \in \mathbb{R}^2 \times \mathbb{S}^1$, where $p_x,p_y$ represent robot position and $\theta$ denotes robot heading angle.

The observation $o \in \Omega$ is derived from LiDAR measurements and geometric descriptors. 
A LiDAR with $G$ beams at resolution $\Delta \theta$ is partitioned into groups of size $g$, with the minimum range from each group retained, yielding $K = G/g$ aggregated features $\{r(\theta_1), \ldots, r(\theta_K)\}$. 
We further extract the $b$ nearest obstacle descriptors $\{(d_i, \phi_i)\}_{i=1}^b$ and the relative goal angle $\phi_g$, forming
\begin{equation}
o = \Big[ r(\theta_1), \ldots, r(\theta_K), \; d_1, \ldots, d_b, \; \phi_1, \ldots, \phi_b, \; \phi_g \Big].
\end{equation}
In our setting, $G=72$, $g=3$, and $b=2$, producing $K=24$ grouped features; together with obstacle and goal descriptors, this yields $o \in \mathbb{R}^{29}$. 
This downsampling compresses dense scans into compact features that preserve proximity cues without explicit obstacle detection, providing an efficient input for policy learning.

\subsection{Problem Statement}
We address the navigation problem of a mobile robot operating in dynamic human environments. 
The goal is to learn a parameterized policy
\begin{equation}
\pi_\theta : \Omega \rightarrow \mathcal{U},
\end{equation}
that maps observations \(o \in \Omega\) to velocity commands \(u \in \mathcal{U}\), enabling safe and adaptive closed-loop navigation. 
To this end, we adopt a teacher–student formulation: the teacher policy evaluates candidate actions using a reward constructed from positive and negative demonstrations together with rule-based specifications, while the student policy \(\pi_\theta\) distills this supervision into a compact controller suitable for real-time deployment.

\section{Proposed Method}
\label{sec:proposed_method}
We present \algoname{} (\algonamelong), a framework for mobile robot navigation in dynamic human environments. 
\algoname{} constructs a reward map by combining density-based rewards inferred from positive and negative demonstrations with rule-based objectives for obstacle avoidance and goal progress. 
This representation encodes human-informed navigation preferences while enforcing safety requirements. 
A sampling-based lookahead controller evaluates candidate rollouts on this reward and selects actions with the highest return, as illustrated in Fig.~\ref{fig:overview}.

The resulting teacher policy benefits from forward simulation to generate safe and adaptive supervision but relies on privileged information through forward simulation of future states. 
To address this, we distill the teacher into a student policy that conditions only on observations. 
The distilled policy retains the adaptability and safety of the teacher and additionally outputs uncertainty estimates that provide indicators of navigation risk in dynamic environments.

\subsection{Teacher Policy via Reward Design}\label{subsec:teacher}
The teacher evaluates short-horizon rollouts generated from sampled velocity commands and selects the first action with the highest return. 
The return is computed from two components: (i) a reward learned from positive and negative demonstrations and (ii) rule-based specifications encoding safety and task objectives. 
This design produces trajectories that reflect demonstrated navigation patterns while maintaining explicit safety margins.

\subsubsection{Density Reward Learning from Positive and Negative Demonstrations}
\label{subsubsec:learning-based}
We learn a reward over state-action pairs tailored to navigation by aligning reward values with the empirical occupancy of demonstrated behavior. 
Let \(\mathcal{S}\) and \(\mathcal{A}\) denote state and action spaces, and write \(x=(s,a)\in\mathcal{X}:=\mathcal{S}\times\mathcal{A}\). 
From demonstrations \(\{x_j\}_{j=1}^{N_D}\subset\mathcal{X}\), define the empirical state--action density \(\hat{\mu}\). 
The reward \(R:\mathcal{X}\to\mathbb{R}\) is obtained by maximizing its expected value under \(\hat{\mu}\),
\begin{equation}
\max_{R}\;\; \langle \hat{\mu}, R\rangle 
\quad \text{subject to} \quad 
\|R\|_{2}\le 1,
\end{equation}\label{eq:roam_obj}
where
\begin{equation}
\langle \hat{\mu}, R\rangle \;=\; \int_{\mathcal{S}\times\mathcal{A}} \hat{\mu}(x)\,R(x)\,dx.
\end{equation}
This formulation is model-free, relying only on demonstrations.

For practical computation and generalization, we represent \(R\) in a reproducing kernel Hilbert space (RKHS) with positive semidefinite kernel \(k\). Using inducing points \(U=\{u_i\}_{i=1}^{N_U}\subset\mathcal{X}\) and coefficients \(\alpha\in\mathbb{R}^{N_U}\),
\begin{equation}
R(x) \;=\; \sum_{i=1}^{N_U}\alpha_i\,k(x,u_i).
\label{eq:roam_rkhs}
\end{equation}
Let \(K_{UU}\in\mathbb{R}^{N_U\times N_U}\) and \(K_{UD}\in\mathbb{R}^{N_U\times N_D}\) have entries \([K_{UU}]_{ij}=k(u_i,u_j)\) and \([K_{UD}]_{ij}=k(u_i,x_j)\). 
To control reward smoothness and improve numerical stability, two quadratic regularization terms are introduced: one weighted by $\lambda$ on the RKHS norm of the function, and another weighted by $\beta$ on the magnitude of the coefficient vector.
With \(\mathbf{1}\in\mathbb{R}^{N_D}\) the all-ones vector and \(\lambda,\beta>0\), we optimize
\begin{equation}
\max_{\alpha\in\mathbb{R}^{N_U}}
\;\;
\frac{1}{N_D}\,\alpha^{\top}K_{UD}\mathbf{1}
\;-\;
\frac{\lambda}{2}\,\alpha^{\top}K_{UU}\alpha
\;-\;
\frac{\beta}{2}\,\alpha^{\top}\alpha,
\label{eq:roam_quad}
\end{equation}
which yields the analytic solution
\begin{equation}
\widehat{\alpha}
\;=\;
\big(\lambda K_{UU}+\beta I_{N_U}\big)^{-1}
\Big(\tfrac{1}{N_D}\,K_{UD}\,\mathbf{1}\Big).
\label{eq:roam_alpha}
\end{equation}

To accommodate demonstrations of positive and negative quality \cite{choi2016robust}, each sample is augmented with a fidelity score \(\gamma\in[-1,1]\), where \(\gamma\approx +1\) denotes desirable behavior, \(\gamma\approx -1\) denotes undesirable behavior, and values near \(0\) indicate uncertain quality.
In our implementation, positive demonstrations are assigned \(\gamma=+1\) and negative demonstrations are assigned \(\gamma=-1\).
We employ a smooth leveraged kernel $k_{\mathrm{SL}}$ that modulates cross-sample similarity by these scores:
\begin{equation}
k_{\mathrm{SL}}\big((x,\gamma),(x',\gamma')\big)
\;=\;
\cos\!\Big(\tfrac{\pi}{2}\,(\gamma-\gamma')\Big)\; k_{\mathrm{PSD}}(x,x'),
\label{eq:roam_sl}
\end{equation}
where \(k_{\mathrm{PSD}}\) is chosen as the widely used squared exponential (SE) kernel
\begin{equation}
k_{\mathrm{SE}}(x,x') \;=\; g^{2}\exp\!\Big(-\tfrac{1}{2l^{2}}\|x-x'\|^{2}\Big),
\label{eq:se_kernel}
\end{equation}
with hyperparameters \(g^{2}\) controlling output scale and \(l^{2}\) the length-scale. 
Inducing points \(U=\{(u_i,\tilde{\gamma}_i)\}_{i=1}^{N_U}\) and demonstrations \(\{(x_j,\gamma_j)\}_{j=1}^{N_D}\) then define \(K_{UU}\) and \(K_{UD}\) via \(k_{\mathrm{SL}}\) for use in \eqref{eq:roam_alpha}. 
This produces a reward estimator explicitly sensitive to the distinction between positive and negative samples.

The resulting formulation yields a density-based reward map that extends traditional occupancy grids by incorporating both physical constraints and demonstrated navigation behaviors. 
It offers three main advantages: (i) \emph{data efficiency}, as rewards are inferred directly from demonstrations without requiring explicit transition dynamics; (ii) \emph{robustness to sample quality}, since the smooth leveraged kernel highlights consistent examples while mitigating the effect of conflicting ones; and (iii) \emph{interpretability and extensibility}, as the reward map is spatially meaningful, reflects demonstrated behaviors, and provides a foundation for integration with complementary priors and control strategies.

\subsubsection{Sampling-Based Lookahead Control}
\label{subsubsec:lookahead}
At each control cycle, candidate actions are evaluated through short-horizon forward simulation. 
For a horizon \(T>0\) and discretization step \(\Delta t\), the rollout of \(u\in\mathcal{A}\) yields a state trajectory
\begin{equation}
\Xi(u) = \{x_\ell(u)\}_{\ell=0}^L, 
\qquad 
L=\big\lfloor T/\Delta t \big\rfloor,
\end{equation}
where \(x_{\ell+1}(u)=\Phi(x_\ell(u),u,\Delta t)\) with \(\Phi\) denoting the one-step integrator.  
Each simulated state is mapped to an observation according to the representation in Sec.~\ref{subsec:state_obs}, producing the observation rollout
\begin{equation}
\xi(u) = \{o_\ell(u)\}_{\ell=0}^L.
\end{equation}

Each rollout \(\xi(u)\) is scored by combining a density-based learned reward from demonstrations (Sec.~\ref{subsubsec:learning-based}) with rule-based priors for goal reaching and obstacle clearance. 
The total score is defined as
\begin{equation}\label{eq:total_reward}
\begin{split}
R_{\text{\algoname{}}}(\xi) 
&= \alpha_{\mathrm{den.}}R_{\mathrm{den.}}(\xi) \\
&\quad + \alpha_{\mathrm{goal}}R_{\mathrm{goal}}(\xi)
+ \alpha_{\mathrm{obs.}}R_{\mathrm{obs.}}(\xi),
\end{split}
\end{equation}
with nonnegative weights \(\alpha_\ast\).

The learned reward $R_{\mathrm{den.}}$ evaluates rollouts using the density-based reward map constructed from positive and negative demonstrations:
\begin{equation}\label{eq:reward_roam}
R_{\mathrm{den.}}(\xi)
=\tfrac{1}{L}\sum_{\ell=1}^L r_{\mathrm{den.}}(o_\ell),
\end{equation}
where \(r_{\mathrm{den.}}(\cdot)\) assigns rewards that capture both physical occupancy and demonstrated navigation preferences, reflecting socially compliant behavior.
As detailed in Sec.~\ref{subsubsec:learning-based}, \(r_{\mathrm{den.}}(\cdot)\) leverages both positive and negative demonstrations, yielding a reward estimator that reflects occupancies and socially compliant behavior.

The goal reward $R_{\mathrm{goal}}$ encourages progress toward the target using the final state $x_{L}(u)$
\begin{equation}\label{eq:reward_goal}
R_{\mathrm{goal}}(\xi) 
= 1 - \tanh\!\left(\tfrac{d_{\mathrm{goal}}}{d_{\mathrm{total}}}\right),
\end{equation}
where \(d_{\mathrm{goal}}\) is the distance from the terminal pose to the goal and \(d_{\mathrm{total}}\) is the initial start-goal distance.

The obstacle reward $R_{\mathrm{obs.}}$ promotes clearance by exploiting nearest-obstacle descriptors embedded in the observations. 
At step \(\ell\), the observation includes the $b$ smallest LiDAR ranges \(\{d_{\ell,1},\ldots,d_{\ell,b}\}\); define per-step clearance as \(d_\ell=\min_i d_{\ell,i}\). 
The obstacle reward is defined as the average per-step clearance over the rollout
\begin{equation}\label{eq:reward_obs}
R_{\mathrm{obs.}}(\xi) = \tanh\!\left(\tfrac{1}{L}\sum_{\ell=1}^L d_\ell\right).
\end{equation}

To achieve a context-aware balance among the three components, the weights \(\alpha_\ast\) are modulated as a function of the remaining goal distance. 
Let 
\begin{equation}
r \;\triangleq\; \mathrm{clip}\!\left(\tfrac{d_{\text{goal}}}{d_{\text{total}}},\,0,\,1\right),
\end{equation}
where \(d_{\text{goal}}\) is the current distance from the rollout endpoint to the goal and \(d_{\text{total}}\) is the initial start--goal distance.  
The coefficients are then defined as
\begin{equation}\label{eq:reward_alpha}
\alpha_{\text{den.}} \;=\; 1+\cos\!\big(\pi(1-r)\big), 
\,\,
\alpha_{\text{goal}} \;=\; 2(1-r),
\,\,
\alpha_{\text{obs}} \;=\; 1.
\end{equation}
This scheme emphasizes the density reward in the early phase, gradually shifts weight toward goal-directed behavior as the robot nears the target, and maintains obstacle avoidance as a constant priority. 
Overall, the adaptive weighting yields smooth transitions between safety, adaptability, and efficiency.

In practice, the controller operates at 10 Hz with \(T=3\) s and \(\Delta t=0.3\) s. 
An exponential moving average with coefficient \(\alpha_{\text{EMA}}=0.5\) is applied to successive velocity commands to reduce abrupt changes and promote stable navigation. 
The candidate action set is constructed by pairing discrete linear and angular velocities: the linear velocity is selected from \(\{0.1,0.2,\ldots,0.8\}\,\text{m/s}\), and the angular velocity is chosen from 15 uniformly spaced values within \([ -0.4\pi,\ 0.4\pi ]\,\text{rad/s}\). 
By combining the density-based learned reward with adaptive weighting of goal and obstacle terms, the lookahead controller achieves context-aware navigation that remains both flexible and safety-oriented in dynamic human environments.

\begin{figure*}[t]
    \centering
    \begin{subfigure}[t]{0.30\textwidth}
        \includegraphics[width=\linewidth]{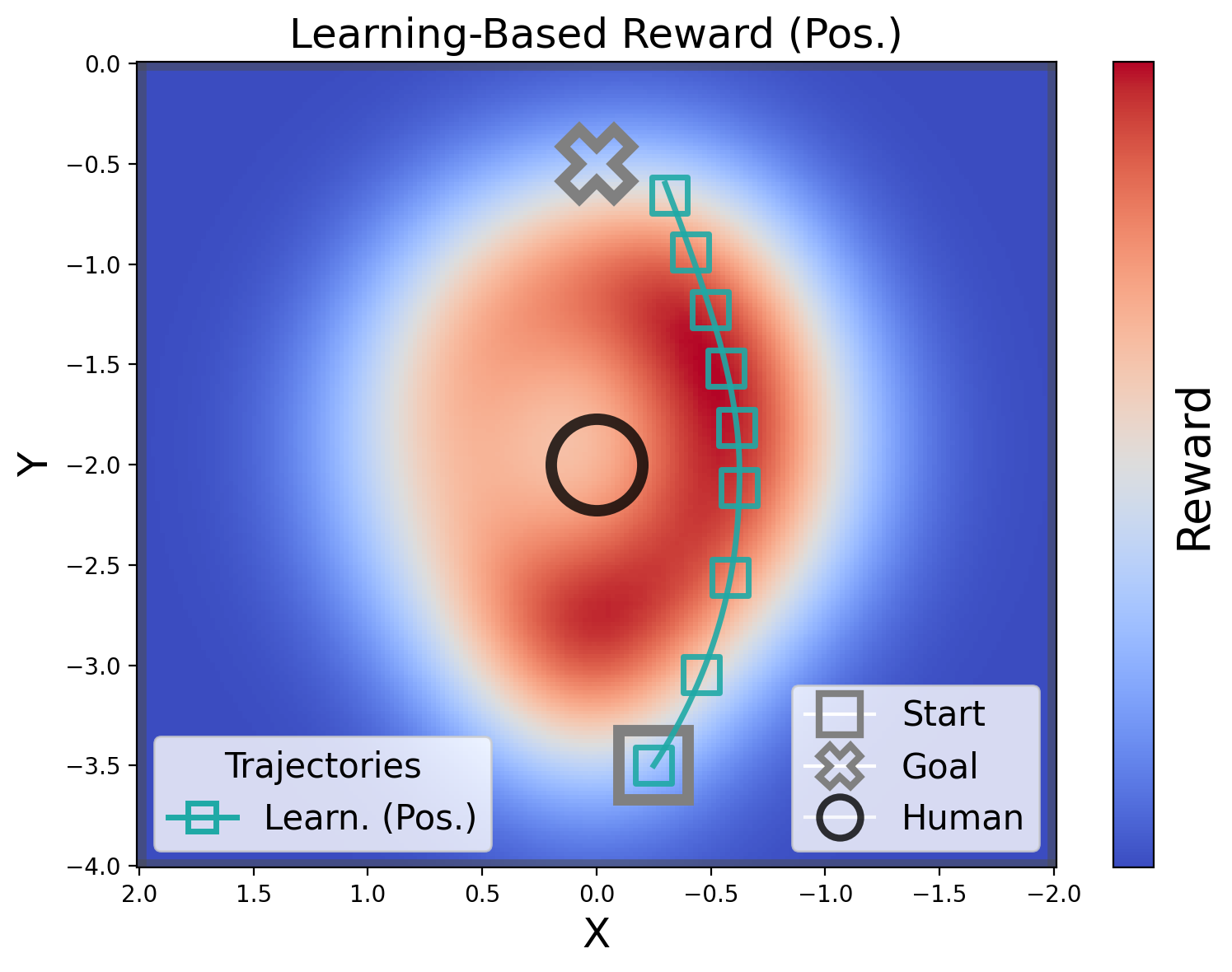}
        \caption{}
        \label{reward_pos}
    \end{subfigure}\hfill
    \begin{subfigure}[t]{0.30\textwidth}
        \includegraphics[width=\linewidth]{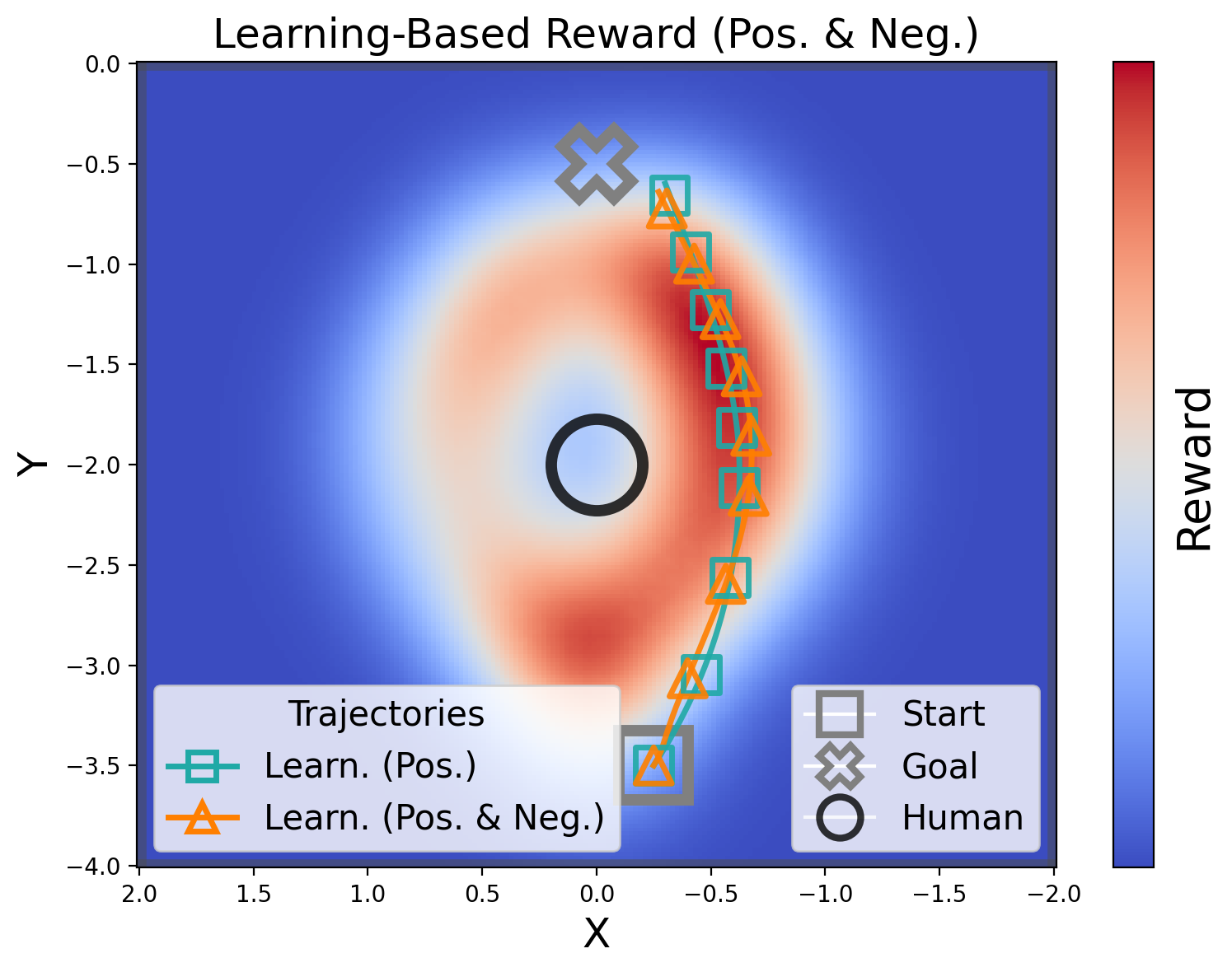}
        \caption{}
        \label{reward_mixed}
    \end{subfigure}\hfill
    \begin{subfigure}[t]{0.30\textwidth}
        \includegraphics[width=\linewidth]{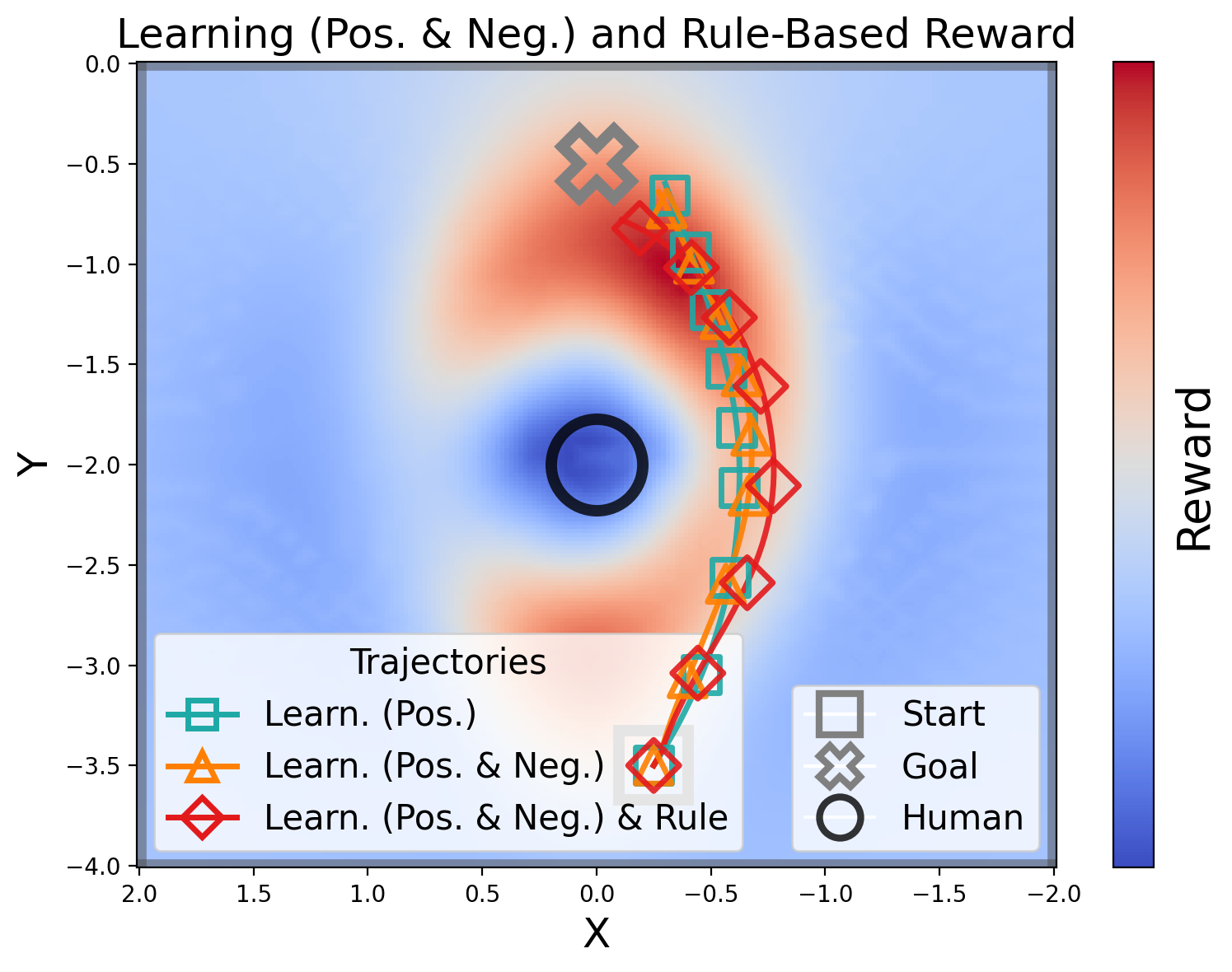}
        \caption{}
        \label{trajectories}
    \end{subfigure}
    \caption{Reward maps and resulting trajectories with synthetic dataset. (a) Learning-based reward map trained with positive samples, highlighting both feasible corridors. (b) Learning-based reward map trained with positive and negative samples, reducing reward near humans. (c) Reward map combining learning-based and rule-based components, yielding trajectories with greater clearance.}
    \label{fig:synthetic}
    \vspace{-10pt}
\end{figure*}

\subsection{Uncertainty-Aware Distillation}
The teacher policy relies on privileged information and short-horizon simulation, which makes it unsuitable for direct deployment under real-time constraints. 
Therefore, we distill the teacher into a compact student policy \(\pi_{\phi}(u\mid o)\) using dataset aggregation~\cite{ross2011reduction}. 
The student is parameterized as a Mixture Density Network (MDN)~\cite{bishop1994mixture}, which models the conditional distribution over velocity commands as a Gaussian mixture:
\begin{equation}
p(u\mid o) \;=\; \sum_{k=1}^{K}\pi_k(o)\,\mathcal{N}\!\big(u\mid \mu_k(o), \Sigma_k(o)\big),
\end{equation}
where the mixture weights \(\pi_k(o)\), means \(\mu_k(o)\), and diagonal covariances \(\Sigma_k(o)\) are predicted by a neural network. 
The network is trained to maximize the likelihood of the teacher’s actions, enabling the student to reproduce expert guidance from  observations.

An additional property of the MDN is that it provides closed-form uncertainty estimates via the law of total variance~\cite{choi2018uncertainty}. 
The predictive covariance decomposes into aleatoric terms, capturing inherent data noise, and epistemic terms, reflecting how each predicted value is different from others:
\begin{equation}
\begin{aligned}
\mathbb{V}(y\mid x)
&=
\mathbb{E}_{k\sim \pi}\!\big[\mathbb{V}(y\mid x,k)\big]
\;+\;
\mathbb{V}_{k\sim \pi}\!\big(\mathbb{E}[y\mid x,k]\big),
\\[0.25em]
&=
\underbrace{\sum_{j=1}^{K}\pi_j(x)\,\Sigma_j(x)}_{\text{\(\Sigma_{\mathrm{alea.}}(x)\)}}
+\underbrace{\sum_{j=1}^{K}\pi_j(x)\,\big\|\mu_j(x)-\mathbb{E}[y\mid x]\big\|_2^{2}}_{\text{\(\Sigma_{\mathrm{epis.}}(x)\)}}.
\end{aligned}
\end{equation}
These uncertainty measures provide informative signals for risk-aware analysis \cite{choi2018uncertainty,oh2022towards}, as elevated values may be associated with closer human–robot interactions. 

\section{Experiments}
In this section, we present a series of experiments to validate the proposed framework.
First, in Sec.~\ref{subsec:synthetic}, a controlled synthetic study examines the contribution of each component, comparing models trained only with positive demonstrations, augmented with negative demonstrations encoding unsafe behaviors, and further combined with rule-based specifications.
This analysis highlights how density-based rewards capture demonstrated preferences, while rule-based terms enforce explicit safety margins and goal achievement.
Subsequently, in Sec.~\ref{subsec:dynamic}, we assess \algoname{} in a realistic elevator co-boarding simulation, where humans are modeled using a social force model, and provide quantitative comparisons against baseline methods.
This scenario is particularly well-suited for evaluation, as it naturally induces dense interactions, intent uncertainty, and social conventions within a confined space, thereby reflecting the core challenges of socially aware navigation.
We then analyze the role of uncertainty-aware distillation in Sec.~\ref{subsubsec:uncertainty}, showing that predictive uncertainty serves as an effective signal for distinguishing safe and risky navigation. 
Finally, Sec.~\ref{subsec:real} reports real-world experiments in elevator scenarios, confirming that \algoname{} achieves safe, efficient, and socially compliant navigation.

\subsection{Synthetic Study in a Static Environment}\label{subsec:synthetic}
We conduct a synthetic study in a planar static setting to illustrate how positive and negative demonstrations shape the learned reward and resulting navigation behavior. 
For interpretability, the reward map is defined over \((x,y)\) coordinates, while trajectories are generated in the full state--action space \((x,y,\theta,v,\omega)\) using the sampling-based lookahead controller. 
Demonstrations were collected through keyboard teleoperation, yielding $446$ positive and $337$ negative samples distributed across left and right corridors around a central human.

Figure~\ref{fig:synthetic} compares reward maps and resulting trajectories under different conditions. 
In Fig.~\ref{fig:synthetic}(a), training with only positive demonstrations highlights both feasible corridors, with a bias toward the right where demonstrations are more frequent. 
In Fig.~\ref{fig:synthetic}(b), negative demonstrations corresponding to human collisions are incorporated. 
This reduces reward values in unsafe regions near the human and shifts the preferred trajectory toward the right corridor. 
In Fig.~\ref{fig:synthetic}(c), the learned reward is further combined with rule-based specifications for obstacle avoidance and goal seeking, resulting in trajectories that preserve clearance around the human while achieving the navigation objective. 
Although negative demonstrations here represent collisions, the same formulation can also encode broader social norms by treating undesired behaviors as negative examples.

\subsection{Dynamic Environments in Elevator Co-Boarding}
\label{subsec:dynamic}
We then evaluate \algoname~in a dynamic setting that captures human motion during elevator co-boarding. 
Humans are simulated with a social force model~\cite{helbing1995social}, and the robot receives LiDAR observations \(o \in \Omega\) and issues velocity commands \(u=[v,w]^T \in \mathcal{U}\). 
Within \algoname{}, a density-based reward is learned from demonstration pairs \((\Omega\times \mathcal{U})\) and combined with rule-based specifications for obstacle avoidance and goal seeking. 
The resulting reward formulation is then used by the sampling-based lookahead controller to generate actions.  

To reflect realistic co-boarding scenarios, we consider two representative human–robot placements: HR-RL (Human-Right, Robot-Left) and HL-RR (Human-Left, Robot-Right). 
Training data were collected via keyboard teleoperation for both scenarios: approximately $732$ positive and $170$ negative demonstrations in the HR-RL setting, and $662$ positive and $194$ negative demonstrations in the HL-RR setting. 
Performance is reported over five random seeds with one hundred trials per seed, using success rate (SR), total time (TT), and path length (PL) as evaluation metrics.

The simulation environment is designed to emulate a realistic elevator setting. 
The global map spans \(4 \times 4\,\text{m}^2\), and the elevator geometry is modeled after a Schindler 6000 unit \cite{schindler6000} with door width of \(1.8\,\text{m}\), cabin width of \(2.5\,\text{m}\), and depth of \(2.7\,\text{m}\). 
The scenario models two humans boarding the elevator while one exits, with the robot initialized at a random position between the two boarding individuals. 
This setup induces encounters near the doorway as the exiting human crosses paths with the robot. 
In the HR-RL setting, humans board on the right side of the elevator and the robot’s goal is on the left; in the HL-RR setting, humans board on the left and the goal is on the right. 
This configuration induces structured interactions at the doorway and requires the robot to navigate through crossing pedestrian flows.

Human motion during elevator co-boarding is modeled using the social-force framework \cite{helbing1995social}, where each agent follows a preferred velocity \(v_{\text{pref}}\) toward its goal while being influenced by repulsive forces from surrounding agents. 
In our setting, the model assumes uncooperative behavior in which humans primarily avoid collisions with other agents, while additional repulsive terms account for interactions with the walls and the robot. 
Following prior implementations \cite{wang2025safe, liu2023intention}, \(v_{\text{pref}}\) is randomly sampled from \(\{0.5, 0.6, 0.7\}\,\text{m/s}\), and each human is modeled with a radius of \(0.6\,\text{m}\). 
This configuration yields realistic yet challenging co-boarding scenarios in which the robot must adapt its navigation strategy to human flows while preserving safety.

For comparison, we evaluate two representative baselines. CVaR-BF~\cite{wang2025safe} is an optimization-based controller that combines Conditional Value-at-Risk (CVaR) with control barrier functions. 
It adaptively adjusts risk levels and expands safety margins through dynamic barrier zones, serving as a safety filter that minimally modifies nominal controls while prioritizing collision avoidance. 
CrowdNav++~\cite{liu2023intention} is a learning-based approach that models human–robot interactions with a spatio-temporal graph and attention, distinguishing robot–human and human–human relations. 
By integrating multi-step human intention prediction into a reinforcement learning framework, CrowdNav++ produces intention-aware and collision-averse navigation behaviors. 

\begin{table}[t]
\centering
\small
\renewcommand{\arraystretch}{1.15}
\caption{Performance comparison in two representative human–robot elevator co-boarding scenarios: HR-RL (Human-Right, Robot-Left) and HL-RR (Human-Left, Robot-Right).}
\label{tab:teacher}

\sisetup{detect-weight=true, table-number-alignment=center}

\begin{tabularx}{\linewidth}{@{} C{1.0cm} Y
  S[table-format=2.1]
  S[table-format=2.2]
  S[table-format=1.2] @{}}
\toprule
\textbf{Scenario} & \textbf{Method} &
\multicolumn{1}{c}{\textbf{SR $\uparrow$ (\%)}} &
\multicolumn{1}{c}{\textbf{TT $\downarrow$ (s)}} &
\multicolumn{1}{c}{\textbf{PL $\downarrow$ (m)}} \\
\midrule
\multirow{3}{*}{HR-RL}
 & Ours                          & 99.4 & 12.24 & 3.74 \\
 & CVaR-BF~\cite{wang2025safe}   & 72.8 & 12.82 & 4.65 \\
 & CrowdNav++~\cite{liu2023intention} & 78.6 & 17.52 & 4.93 \\
\midrule
\multirow{3}{*}{HL-RR}
 & Ours                          & 99.6 & 12.94 & 3.88 \\
 & CVaR-BF~\cite{wang2025safe}   & 71.4 & 12.82 & 4.75 \\
 & CrowdNav++~\cite{liu2023intention} & 64.8 &  9.71 & 4.63 \\
\bottomrule
\end{tabularx}
\vspace{-10pt}
\end{table}

\subsubsection{Navigation Performance}
Table~\ref{tab:teacher} reports navigation performance in the HR-RL and HL-RR elevator co-boarding scenarios. 
Across both settings, the proposed method achieves success rates exceeding 99\% while maintaining low total time (TT) and path length (PL), indicating reliable and efficient navigation in dynamic human environments. 
Compared to CVaR-BF~\cite{wang2025safe} and CrowdNav++~\cite{liu2023intention}, our approach achieves substantially higher success while preserving efficiency, highlighting the benefit of integrating demonstration-driven rewards with rule-based safety specifications. 

Table~\ref{tab:ablation} presents ablation results isolating the contributions of individual components. 
Removing the density-based reward leads to the largest degradation in success rate, underscoring the importance of demonstration-aligned rewards for resolving congested interactions at the doorway. 
Excluding the obstacle prior reduces total time but decreases success, while excluding the goal prior increases total time due to less directed motion. 
These trends indicate that both learned rewards and rule-based priors play complementary roles, and their combination is essential for balancing adaptability with explicit safety margins. 

\begin{table}[t]
\centering
\small
\renewcommand{\arraystretch}{1.15}
\caption{Ablation study of reward terms in two representative human–robot elevator co-boarding scenarios. 
Each row indicates which reward terms are active (\cmark) or removed (\xmark).}
\label{tab:ablation}
\begingroup
\sisetup{detect-weight = true, table-number-alignment = center}

\resizebox{\linewidth}{!}{%
\begin{tabular}{%
  @{} C{1.2cm}
      *{3}{C{0.7cm}}
      S[table-format=2.1, table-text-alignment=center]
      S[table-format=2.2, table-text-alignment=center]
      S[table-format=1.2, table-text-alignment=center]
  @{}}
\toprule
\multirow{2}{*}[-\dimexpr\aboverulesep/2+\belowrulesep/2\relax]{\textbf{Scenario}} &
\multicolumn{3}{c}{\textbf{Reward terms}} &
\multicolumn{1}{c}{\multirow{2}{*}[-\dimexpr\aboverulesep/2+\belowrulesep/2\relax]{\textbf{SR $\uparrow$ (\%)}}} &
\multicolumn{1}{c}{\multirow{2}{*}[-\dimexpr\aboverulesep/2+\belowrulesep/2\relax]{\textbf{TT $\downarrow$ (s)}}} &
\multicolumn{1}{c}{\multirow{2}{*}[-\dimexpr\aboverulesep/2+\belowrulesep/2\relax]{\textbf{PL $\downarrow$ (m)}}} \\
\cmidrule(lr){2-4}
& \textbf{den.} & \textbf{goal} & \textbf{obs.} & & & \\
\midrule
\multirow{3}{*}[-\dimexpr\aboverulesep/2+\belowrulesep/2\relax]{HR--RL}
 & \xmark & \cmark & \cmark & 83.0 & 13.61 & 4.55 \\
 & \cmark & \xmark & \cmark & 99.4 & 15.70 & 3.79 \\
 & \cmark & \cmark & \xmark & 94.8 & 11.64 & 3.72 \\
\midrule
\multirow{3}{*}[-\dimexpr\aboverulesep/2+\belowrulesep/2\relax]{HL--RR}
 & \xmark & \cmark & \cmark & 79.8 & 13.37 & 4.54 \\
 & \cmark & \xmark & \cmark & 99.4 & 16.48 & 3.92 \\
 & \cmark & \cmark & \xmark & 98.8 & 12.33 & 3.86 \\  
\bottomrule
\end{tabular}%
}
\endgroup
\vspace{-6pt}
\end{table}

\subsubsection{Uncertainty-Aware Distillation}
\label{subsubsec:uncertainty}
\begin{figure*}[!t]
    \centering
    \includegraphics[width=1.85\columnwidth]{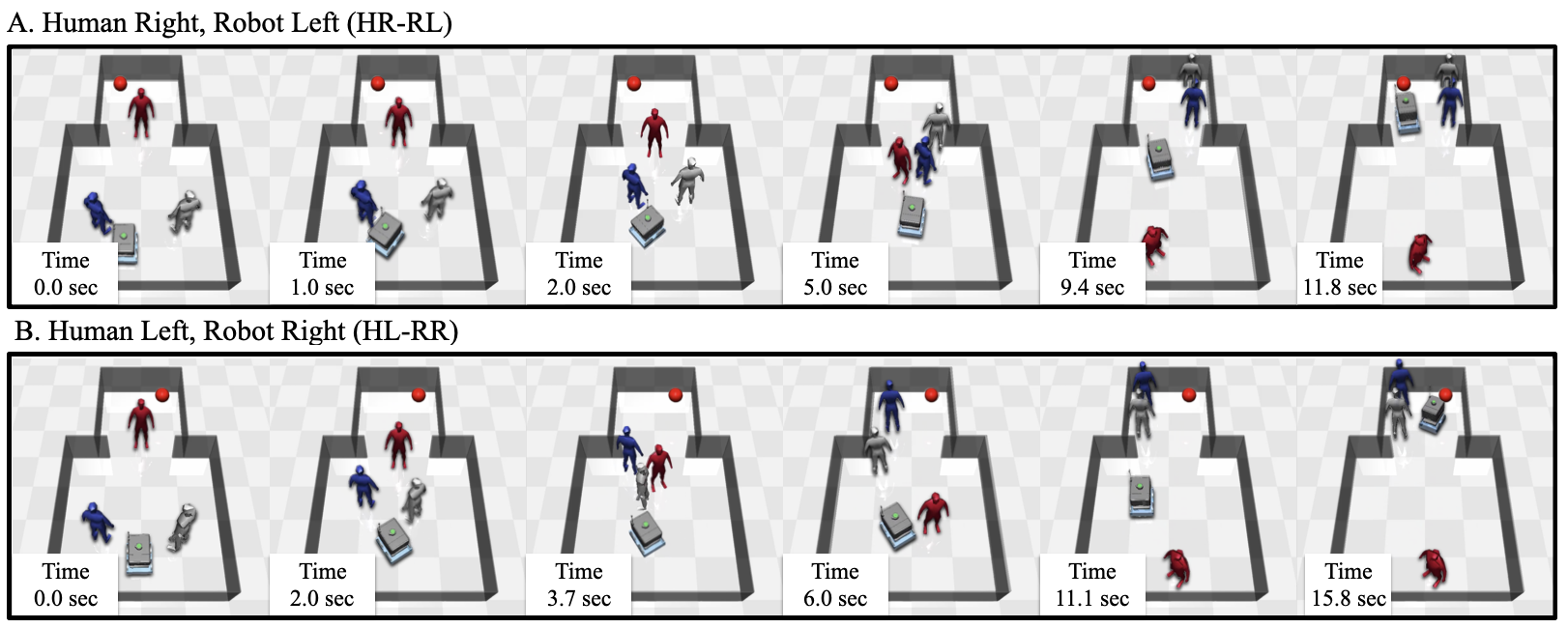}
    \caption{
    Simulation snapshots of elevator co-boarding scenarios. 
    (A) HR-RL: Human on the right, Robot on the left. 
    (B) HL-RR: Human on the left, Robot on the right.
    }
    \label{fig:sim_snapshot}
    \vspace{-4mm}
\end{figure*}

We distill the teacher policy into a mixture density network (MDN) to capture navigation behaviors. 
Data collection follows the DAgger protocol~\cite{ross2011reduction}, beginning with 50 expert episodes and continuing with 10 rounds of 50 episodes each, where the student executes the policy while the teacher provides action labels. 
Training uses a batch size of 256 and a learning rate of \(10^{-3}\). 
The MDN comprises two hidden layers (128 units, ReLU, dropout 0.1) and a Gaussian-mixture output head with \(K=10\) components that predicts mixture weights, means, and variances. 
For comparison, we also distill into a standard MLP with the same hidden layer configuration. 
As in Table~\ref{tab:student}, the MDN achieves higher success rates than the MLP in both HR-RL and HL-RR scenarios, while also yielding shorter travel times and path lengths. 
Representative trajectories of the distilled policy are shown in Fig.~\ref{fig:sim_snapshot}, demonstrating navigation performance in dynamic elevator co-boarding environments.

Beyond overall performance, we further assess whether the distilled policy can provide signals of risk through predictive uncertainty. 
Evaluation data are partitioned into safe and risky frames according to the minimum human-robot distance: frames with clearance greater than \(0.8\,\text{m}\) are labeled safe, while those below this threshold are labeled risky. 
In the HR-RL setting, this results in 62{,}374 safe frames and 5{,}467 risky frames, and in the HL-RR setting, 67{,}806 safe frames and 3{,}245 risky frames. 
Across both scenarios, epistemic and aleatoric uncertainty values are consistently higher in risky frames, indicating that the MDN provides informative signals about potential hazards. 
Such estimates can guide conservative fallback strategies such as slowing, yielding, or reverting to rule-based control~\cite{choi2018uncertainty}, thereby complementing the efficiency of the distilled policy with explicit safety awareness.

\subsection{Real-World Demonstrations}
\label{subsec:real}
\begin{figure*}[!t]
    \centering
    \includegraphics[width=1.90\columnwidth]{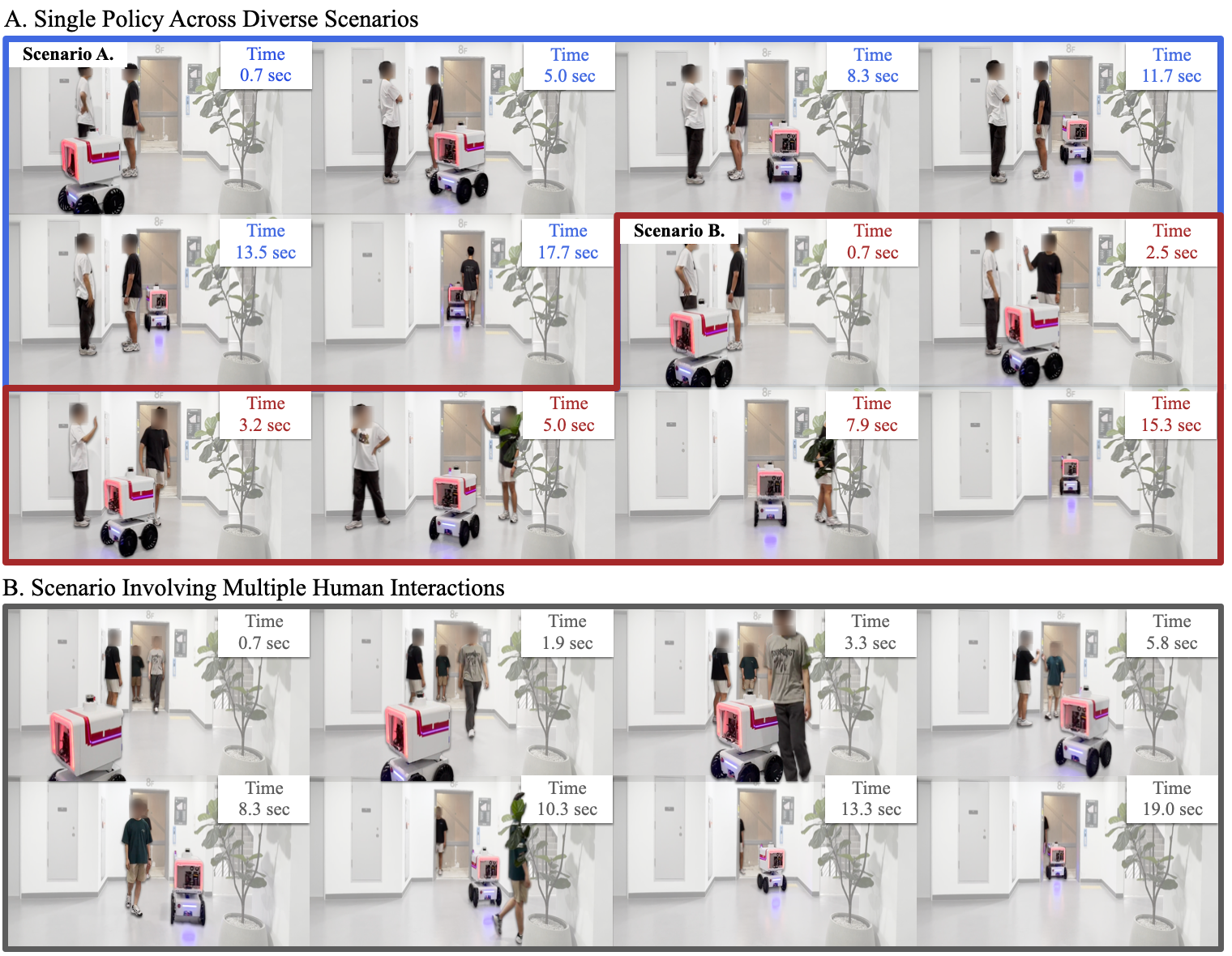}
    \caption{
    Real-world elevator co-boarding experiments with human participants. 
    (A) Single policy across diverse scenarios. 
    (B) Scenario involving multiple human interactions. 
    }
    \label{fig:realworld}
    \vspace{-20pt}
\end{figure*}

To demonstrate the practicality of the proposed framework, we conducted real-world experiments in two representative elevator co-boarding scenarios with human participants, as illustrated in Fig.~\ref{fig:realworld}. 
A four-wheel mobile robot executed control at 10 Hz, with policy observations derived from 3D LiDAR data converted into 2D laser scans. 

In the first scenario, a single policy was deployed across varied co-boarding situations, adapting to differences in human motion. 
In the second scenario, the robot navigated encounters involving multiple humans and managed the interactions effectively. 
These trials demonstrate that the proposed framework extends beyond simulation and supports reliable operation in real-world human–robot interaction settings.

\begin{table}[t]
\centering
\small
\renewcommand{\arraystretch}{1.15}
\caption{Comparison of distillation performance from the teacher policy across different network architectures.}
\label{tab:student}

\sisetup{detect-weight=true, table-number-alignment=center}

\begin{tabularx}{\linewidth}{@{} C{1cm} Y
  S[table-format=3.1]
  S[table-format=2.2]
  S[table-format=1.2] @{}}
\toprule
\textbf{Scenario} & \textbf{Method} &
\textbf{SR $\uparrow$~(\%)} &
\textbf{TT $\downarrow$~(s)} &
\textbf{PL $\downarrow$~(m)} \\
\midrule
\multirow{2}{*}{HR-RL}
 & \algoname{}-MDN & 98.0 & 13.67 & 4.62 \\
 & \algoname{}-MLP & 95.8 & 15.02 & 6.49 \\
\midrule
\multirow{2}{*}{HL-RR}
 & \algoname{}-MDN & 100.0  & 14.16 & 4.70 \\
 & \algoname{}-MLP & 98.8 & 15.29 & 4.78 \\
\bottomrule
\end{tabularx}
\vspace{-10pt}
\end{table}

\subsection{Limitations}
While the proposed framework demonstrates strong performance in both simulation and real-world experiments, certain limitations remain. 
First, the simulation of human motion is based on a social-force model, which provides structured scenarios but does not fully capture the variability of real pedestrian behaviors. 
Second, although the reward learning formulation supports a continuous fidelity variable \(\gamma \in [-1,1]\) to represent a spectrum of demonstration quality, in practice we restricted \(\gamma\) to binary labels for positive and negative examples. 
This simplification improves training efficiency but may limit the ability to capture more nuanced variations in demonstration quality.

\section{Conclusion}
This paper presented a navigation framework that combines density-based reward learning from positive and negative demonstrations with rule-based specifications for obstacle avoidance and goal seeking. 
The teacher policy evaluates short-horizon rollouts under this reward formulation, and a student policy is distilled for real-time deployment while retaining risk-awareness through predictive uncertainty. 
Experiments in synthetic and dynamic elevator co-boarding scenarios showed that the proposed approach improves success rates and efficiency compared to baselines, while real-world trials confirmed its feasibility on a mobile robot operating alongside humans. 
These results demonstrate the effectiveness of integrating data-driven rewards with explicit safety specifications for socially compliant robot navigation.

\begin{table}[t]
\centering
\small
\setlength{\tabcolsep}{6pt}
\renewcommand{\arraystretch}{1.2}
\caption{Mean uncertainty values for safe–risky frames.}
\label{tab:mean_uncertainty}
\begin{tabular}{
  @{} l
  c c  c c   
  @{}
}
\toprule
\multirow{2}{*}{\textbf{Scenario}} &
\multicolumn{2}{c}{\textbf{Safe}} & \multicolumn{2}{c}{\textbf{Risky}} \\
\cmidrule(lr){2-3} \cmidrule(lr){4-5}
& \textbf{Epi.} & \textbf{Ale.} & \textbf{Epi.} & \textbf{Ale.} \\
\midrule
HR-RL & 0.290 & 0.310 & 0.382 & 0.407 \\
HL-RR & 0.342 & 0.270 & 0.523 & 0.426 \\
\bottomrule
\vspace{-8mm}
\end{tabular}
\end{table}

\bibliographystyle{IEEEtran}
\bibliography{references}

\end{document}